\documentclass{article}
\usepackage{spconf,amsmath,graphicx}

\usepackage{epsfig}
\usepackage{amsfonts}
\usepackage{multirow}
\usepackage{url}

\usepackage[utf8]{inputenc}
\usepackage{cleveref}
\crefname{section}{§}{§§}
\Crefname{section}{§}{§§}
\usepackage{times}
\usepackage{latexsym}

\usepackage[T1]{fontenc}

\usepackage[utf8]{inputenc}

\usepackage{microtype}

\usepackage{algorithm2e}
\RestyleAlgo{ruled}

\usepackage{ragged2e}
\usepackage{amssymb}

\usepackage{graphicx}
\graphicspath{ {./images/} }

\usepackage{subfig}

\usepackage{amsmath}
\usepackage{bm}

\usepackage{longtable}
\usepackage{booktabs} 
\usepackage{multirow}

\newcommand{\docred}{DocRED}
\newcommand{\docredfe}{DocRED-FE}

\title{DocRED-FE: A Document-level Fine-grained Entity and Relation Extraction Dataset}
\name{Hongbo Wang$^{}\sthanks{\quad indicates equal contribution}$, Weimin Xiong$^{*}$, Yifan Song, Dawei Zhu, Yu Xia, Sujian Li$^{}$\sthanks{\quad Corresponding author}}

\address{National Key Laboratory for Multimedia Information Processing, Peking University \\
whb@stu.pku.edu.cn, \{wmxiong, yfsong, dwzhu, yuxia, lisujian\}@pku.edu.cn}

\begin{document}
%
\maketitle
\begin{abstract}
Joint entity and relation extraction (JERE) is one of the most important tasks in information extraction.
However, most existing works focus on sentence-level coarse-grained JERE, which have limitations in real-world scenarios.
In this paper, we construct a large-scale document-level fine-grained JERE dataset \docredfe{}, which improves \docred{} with \textbf{F}ine-Grained \textbf{E}ntity Type.
Specifically, we redesign a hierarchical entity type schema including 11 coarse-grained types and 119 fine-grained types, and then re-annotate  \docred{} manually according to this schema. Through comprehensive experiments we find that: 
(1) DocRED-FE is challenging to existing JERE models; 
(2) Our fine-grained entity types promote relation classification.
We make DocRED-FE with instruction and the code for our baselines publicly available at \url{https://github.com/PKU-TANGENT/DOCRED-FE}.
\end{abstract}

\begin{keywords}
 Joint Entity and Relation Extraction, Information Extracion, Fine-Grained Entity Types
\end{keywords}
%

\begin{figure}[t]
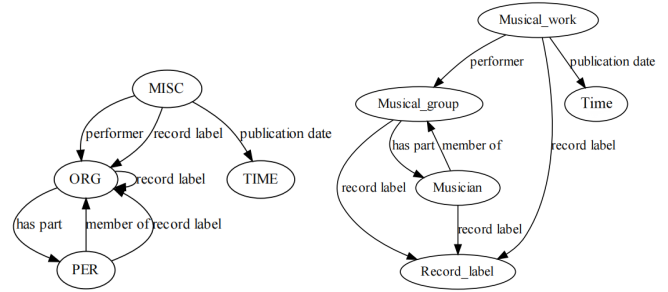

    \centering
    \subfloat{\includegraphics[width = 0.9\linewidth]{case.pdf}}
    \quad
    \setlength{\lineskip}{-0.5em}  
    \subfloat{\includegraphics[width = 0.5\linewidth]{schema_before.pdf}}
    \subfloat{\includegraphics[width = 0.5\linewidth]{schema_after.pdf}}
    \caption{A document case of \docredfe{}, with the corresponding origin (left) and new (right) schema. Only highlights the first mention of each entity.}
    \label{fig:schema}
    \vspace{-0.3cm}
\end{figure}

\section{Introduction}
\label{sec:intro}



The goal of joint entity and relation extraction (JERE) is to identify named entities and their relations from unstructured text. 
It is an essential problem in information extraction (IE) since it is critical to constructing knowledge graph. 
Recently, various works on JERE, including pipelined and end-to-end approaches, have achieved remarkable success \cite{bekoulis_2018_joint, wang_2020_pretraining, zhong_2021_frustratingly}.

Despite these successful efforts, most existing JERE works are conducted on sentence-level coarse-grained datasets, \textsl{e.g.}, ACE04/05~\cite{ace}, CoNLL04~\cite{conll}, SciERC~\cite{scierc}, TACRED~\cite{zhang2017position}, FewRel~\cite{han2018fewrel}.
However, in real-world application scenarios, the model usually needs to understand the semantics of document-level texts and extract more precise information.

The research on document-level fine-grained entity and relation extraction requires a large-scale annotated dataset for both training and evaluation.
It is usually expensive and time consuming to construct such a large-scale dataset from scratch.
Therefore, we choose to re-annotate a popular document-level relation extraction dataset, \docred{}~\cite{yao2019docred}.
There are 96 fine-grained relation types in \docred{}, while the entities are divided into only 6 types and are recognized by automatic tools.


In this paper, we propose \docredfe{}, a large-scale human-annotated fined-grained entity and relation extraction dataset based on \docred{}. 
Specifically, we first re-design the entity type schema to a hierarchical structure including 11 coarse-grained types and 119 fine-grained types.
Combined with the 96 fine-grained relation types in original \docred{}, \docredfe{} contains richer contextual information with a finer granularity.
We illustrate an example document with its corresponding schema in Fig.~\ref{fig:schema}.
Compared to the coarse schema in \docred{}, our newly designed entity schema is more precise and expressive with most of the self-pointing relation edges dissolved, such as \textit{ORG-record\_label-ORG}.

To assess the JERE benchmark on \docredfe{}, we adopt several representative document-level JERE baselines, including a pipeline model based on BERT~\cite{devlin2018bert} and an end-to-end model based on BART~\cite{lewis-etal-2020-bart}.
Experimental results show that document-level fine-grained JERE poses a great challenge to current models and remains an open problem. Furthermore, we also confirm that fine-grained entity information can improve the performance on relation classification (RC). We believe our analysis can help to design more powerful models.

\section{Dataset Construction}
\label{sec:Dataset Construction}

\docredfe{} is built based on \docred{}, a large-scale crowd-sourced dataset from Wikipedia and Wikidata.
\docred{} is mainly built for relation extraction and has only 6 coarse entity types. 
Thus, we develop \docredfe{} via refining the entity type schema and re-annotating the entities in \docred{}.

\subsection{Schema Design}
In this section, we introduce our \emph{bottom-up} data-driven annotation approach.
We aims to establish a schema which can reflect the characteristics of entities in texts properly, and make the frequency of each label in schema as balanced as possible. Therefore, controlling the granularity and dividing boundaries of the schema is particularly critical. 
Formally, we design the entity schema in the following three steps:

\noindent \textbf{Stage 1: Entity Linking.}
For each entity in the dataset, we link it to Wikidata to get its types. For example, for the entity \emph{China}, we get its candidate types as \emph{ {Country, State, Socialist State}}. Then we statistically compute the frequency of each type and filter about top-100 of them to constitute our initial fine-grained candidate set. 

\noindent \textbf{Stage 2: Combining with Existing Schema.}
As for the FIGER~\cite{ling2012fine} and Few-NERD~\cite{ding-etal-2021-nerd}, they are both general encyclopedia datasets. Considering their homology with Doc-RED, we combine their entity schema with our initial candidate set. In the process of combination, both coarse-grained and fine-garined types are adjusted. For example, we prefer to make the initial fine-grained type \emph{GPE} in Few-NERD to be coarse-grained one, while for the types \emph{Military\_operation}, \emph{War}, \emph{Battle} that have similar meaning, we only keep the first one remained. After the prior two stages, a preliminary hierarchy schema is built.


\noindent \textbf{Stage 3: Iteratively Exploratory Annotation and Refinement.}
To validate whether this schema is suitable, we carry out a few rounds of exploratory annotation, and refine the schema according to the feedback by the annotators after each round. 
For example, we subdivide the type \emph{GPE} into more subtypes not only \emph{Country, City}, but also \emph{Continent, State\&Province} since entities with type GPE account for a large proportion. We also merge type \emph{Actor} and \emph{Director} into \emph{Actor\&Director} since they are always used to label the same person and hard to distinguish. Consequently, the finalized schema includes 11 coarse-grained types and 119 fine-grained types is established.


\subsection{Human Annotation}

In this stage, we perform human re-annotation on the entities according to our new schema. To ease the annotation burden, we ask annotators to assign one single type that best suits the entity according to its contexts. For example, sometimes \emph{World War II} represents a book but not the war, so it should be tagged by \emph{Written\_work} rather than \emph{Military\_operation}. Sometimes a person is a \emph{Soldier}, \emph{Politician} and \emph{Artist}, we ask the annotators to select the most prominent type depending on the contexts.

Annotators consist of about 20 experts that work on NLP research. They have linguistic knowledge and are instructed with detailed and formal annotation rules. To ease the annotation, the description of each entity on Wikipedia and its structured information on Wikidata are provided to annotators as references. We ensure that all the annotators are fairly compensated by market price according to their workload. The dataset is randomly divided and delivered to different annotators. To ensure the quality of our dataset, we conduct consistency checking 
via calculating the Cohen’s Kappa Score~\cite{cohen1960coefficient}, and the result is 68.58\%, which indicates a relatively high degree of consistency. 

\begin{table}[t]
\caption{Comparison to well-known RE/JERE datasets (Doc.: document, Tok.: token, Ent.: entity, ET.: entity type, Rel.: relation, RT.: relation type).}
\tiny  
\centering
\footnotesize
\setlength\tabcolsep{2pt}
\begin{tabular}{c c c c c c c}
\toprule
\textbf{Dataset} & \textbf{\# Doc.} & \textbf{\# Tok.} & \textbf{\# Ent.} & \textbf{\# ET.} & \textbf{\# Rel.} & \textbf{\# RT.}\\
\midrule

{ACE2005} & {-} & {259k} & {37,622} & {51} & {7,786} & {18}\\
{TACRED} & {-} & {3,866k} & {212,528} & {17} & {106,264} & {41}\\
{SciERC} & {-} & {65,334} & {1,015} & {6} & {2,687} & {7}\\
{FewRel} & {-} & {1,397k} & {112,000} & {-} & {56,000} & {100}\\
{DWIE} & {802} & {501k} & {23,130} & {311} & {21,749} & {65}\\
{DocRED} & {5,053} & {1,002k} & {98,610} & {6} & {56,354} & {96}\\
\midrule
\textbf{\docredfe} & \textbf{2,596} & \textbf{516k} & \textbf{50,549} & \textbf{119} & \textbf{32,366} & \textbf{96}\\

\bottomrule
\end{tabular}
\label{tab:statistic}
\end{table}

\section{Data Analysis}
In this section, we analyze our dataset from various aspects to have a deeper understanding.


\subsection{Comparison with Related Datasets}
Without the test part, \docredfe{} consists of 65\% of the train part and whole dev part of the DocRED. Table ~\ref{tab:statistic} shows statistics of our dataset and some well-known RE/JERE datasets. Except DWIE~\cite{zaporojets2021dwie}, all of above are only fine-grained in either entity or relation. As for DWIE, we do not have enough statistics on entity information since they do not provide a complete entity schema. To the best of our knowledge, it has fewer documents and tokens but more complicated entity types, which make NER difficult due to insufficient training. 

\subsection{Entity Distribution}

\begin{figure}[tbp]
    \vspace{-0.3cm}
    \includegraphics[width = 1\linewidth]{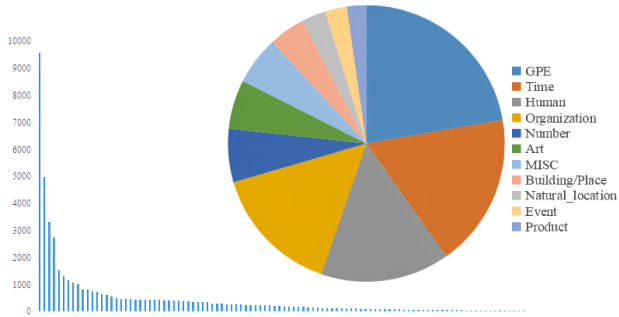}
    \caption{Statistics on first-level (pie chart) and second-level (bar chart) entity type.}
    \label{fig:statisticChart}
\end{figure}

In Figure~\ref{fig:statisticChart}, the pie chart shows the entity distribution of first-level entity type while the bar chart shows that of the second-level. For the bar chart, the top-5 types are \emph{Time}, \emph{Country}, \emph{Number}, \emph{City} and \emph{Musician}, the last-5 are \emph{Chemical\_and\_biological}, \emph{Natural\_phenomenon}, \emph{Dam}, \emph{Disease} and \emph{Sports\_season}.
Long-tail phenomenon still exists, but the downtrend is relatively flat after the top-5.

\subsection{Schema for Each Document}
After our transformation from the original coarse schema to the finer one, the triplets in our schema also differ from  \docred{} due to the variety of the head and tail entity type as shown in Figure~\ref{fig:schema}. Compared with the latter, there are following optimizations:
(1) Diversity of the entity type makes the relation distribution more even, amount of self-pointing edge descends, and overall density tends to be more reasonable.
(2) Naming consistency of entity and relation makes the meaning clear and unambiguous, domain information like music becomes prominent.
(3) The rich entity information can provide more significant hints for RC tasks. To measure the gain quantitatively, we compute the information gain $\Delta H$ and the information gain ratio $\Delta H_{ratio}$ comparing to the original \docred{}. 
The information entropy means average weighted information entropy of the triplet set computed by each entity type pair~\cite{thomas2006elements}, the $\Delta H$ means the value of decline of the information entropy between \docred{} and \docredfe{}, the result is 1.09. The $\Delta H_{ratio}$ is 65.67\%, which means the $\Delta H$ relative to the information entropy of \docred{}. It theoretically indicating that fine-grained entity types can benefit RC.

\section{Experiments}


We conduct two experiments using the proposed \docredfe{} dataset. Firstly, in Section 4.1, we conduct a JERE experiment on the \docred{} and \docredfe{} to show that \docredfe{} provides a more difficult benchmark. Secondly, in Section 4.2, we show that the \docredfe{} can help the RC by providing double-level and fine-grained entity type information.

For the JERE task, we adopt two JERE models as baselines of our dataset, including a pipelined model JEREX~\cite{eberts-ulges-2021-end} and an end-to-end generation model REBEL~\cite{huguet-cabot-navigli-2021-rebel-relation}. JEREX links entity recognition and relation classification tasks by feeding the results of entity recognition to the subsequent relation classification. REBEL employs BART-large as the base model and translates a raw input dataset into a set of triplets that can be uniquely decoded into entity-relation pairs. 

For the RC task, we adopt several representative RC models, including CNN, LSTM, BiLSTM proposed from \docred{}, GAIN~\cite{zeng2020double}, JEREX and SSAN~\cite{xu-etal-2021}.

For both experiments, we train the model using the same part of \docred{} and \docredfe{}. As mentioned above, we use 21k entity-relation pair instances as the train part and 10k instances as the validation part. We use the default settings mentioned in the references and fine-tune the model until the loss function converges and report the results on the validation part.

\begin{table}[t]
\caption{Performance of different JERE models on \docred{} and \docredfe{} (relations).}
\footnotesize
\centering
\begin{tabular}{c c c c}
\toprule
\multirow{2.5}{*}{\textbf{Model}} & \multicolumn{3}{c}{\textbf{DocRED}} \\ \cmidrule(r){2-4}
\multirow{2}{*}{} & \textbf{RE F1 (strict)} & \textbf{RE F1 (relaxed)} & \textbf{NER F1}\\
\midrule
JEREX & {40.26} & {40.62} & {80.25} \\
REBEL & {45.29} & {46.45} & {-} \\ \midrule \midrule
\multirow{2.5}{*}{\textbf{Model}} & \multicolumn{3}{c}{\textbf{\docredfe}} \\ \cmidrule(r){2-4}
\multirow{2}{*}{} & \textbf{RE F1 (strict)} & \textbf{RE F1 (relaxed)} & \textbf{NER F1}\\
\midrule
JEREX & {31.52} & {39.02} & {67.74}\\
REBEL & {37.51} & {45.94} & {-}\\
\bottomrule
\end{tabular}
\label{tab:RE}
\end{table}

\subsection{Joint Entity and Relation Extraction}
\label{exp:jere}
In this experiment, we follow previous work~\cite{eberts-ulges-2021-end,taille_2020_let} and use two modes of F1 score. The strict mode counts a relation as correct if and only if its mention span, coreference resolution, entity type and relation type are correct. The relaxed mode focuses on the relation accuracy and ignores the correctness of the entity type. 
When acquiring the golden entity type during pipelined process, 6 entity types in \docred{} are replaced by 119 entity types when testing \docredfe{}.

As shown in Table~\ref{tab:RE}, both the strict and relaxed F1 of a model trained on \docredfe{} are lower than those trained on \docred{}. Strict F1 declines as expected due to the more complex entity types. The reason why relax F1 become slightly lower is that although fine-grained entities provide more information, incorrect entity predictions may have a negative impact on relation classification and degrade the whole JERE performance. The result suggests that \docredfe{} presents great challenges to current models.



\subsection{Relation Classification}
While JERE task on \docredfe{} poses a challenge for existing models, we want to find how our double-level entity types help the RC part of JERE task and provide some reasonable methods. 
Therefore, we utilize different levels of entity information to build the following three methods:

1.Only use the fine-grained entity type id.

2.Use the fine-grained entity type id and its semantic representation.

3.Use the fine-grained entity type id and corresponding coarse-grained entity type id.


\noindent \textbf{Only Fine-Grained Types}
Similar to Section \ref{exp:jere}, for the entity type embedding part in every models, we replace the original coarse-grained types id embedding by our fine-grained types id embedding. As shown in Table~\ref{tab:RC}, fine-grained entities add more information to relation classifier and constrain the appropriate relation types. This helps the model to classify relations better.


\noindent \textbf{Fine-Grained Types with Its Semantic Representation}
While the entity type id provides necessary information for relation classification, the entity type semantic information could also be utilized. From this perspective, we use BERT to encode each entity type such as \emph{Musician}, \emph{Country}, \emph{Military\_operation} to get its 768-dim semantic embedding $e_s$.
In the way of aggregating, we use concatenation or directly adding depending on different models.

For JEREX, we use MLP to map the $e_s$ into target embedding shape, and concatenate the semantic embedding $e_s$, fine-grained type id embedding $e_f$ with the entity-pair embedding $e_{en}$ before feeding to the final relation classifier $RC$. This process can be expressed as the following formula.
\begin{equation}
    Score = \mathbb{RC}[concat(e_{en}, e_f, e_s)]
\end{equation}

For BiLSTM, GAIN, SSAN, we use MLP to map the $e_s$ into the same shape with word embedding $e_w$, so that we can add these embeddings directly in the embedding block which is in the front of the model. By the way, we add \emph{"Padding"} type for non-entity token intentionally. The token embedding $E_t$ of SSAN compute as
\begin{equation}
    E_t = e_w + e_f + e_s * \alpha
\end{equation}
As shown in Table~\ref{tab:DLESE}, model with added semantic information performs better than that without it. F1 of BiLSTM using semantic information improves from 49.28 to 50.26, indicating that adding semantic information in the \docredfe{} dataset is helpful to relation classification.

\noindent \textbf{Both Fine-Grained and Coarse-Grained Types}
Considering information in the hierarchical structure of entity schema of \docredfe{}, we further integrate new coarse-grained type based on first method. For simplicity, we train a coarse-grained type id embedding layer, project the new coarse-grained type id to type embedding $e_c$, and make fusion like the second method.
For BiLSTM, GAIN and JEREX, concatenation operation as
\begin{equation}
    Score = \mathbb{RC}[concat(e_{en}, e_f, e_c)]
\end{equation}
For SSAN, the token embedding $E_t$ compute as
\begin{equation}
    E_t = e_w + (e_f + e_c) * \gamma
\end{equation}

The second and fourth columns of Table~\ref{tab:DLESE} show that by adding double-level entity type information, RC models could utilize coarse-grained and fine-grained entity type jointly, and perform better than only using fine-grained entity type. 
That means, the hierarchical entity types could be leveraged in designing  RC models.


\begin{table}[t]
\caption{Infusing different granularity of entity information into RC models. "Coarse-grained Type" represent the entity type in \docred{}. We report the F1 score on dev set.}
\footnotesize
\centering
\begin{tabular}{c c c}
\toprule
\textbf{Model} & \textbf{Coarse-grained Type} & \textbf{Fine-grained Type}\\ \midrule
CNN & {43.90} & {44.59} \\
LSTM & {47.72} & {49.34} \\
BiLSTM & {48.64} & {49.28}\\
GAIN & {58.09} & {58.35}\\
JEREX & {57.99} & {58.19}\\
SSAN & {58.10} & {58.24} \\
\bottomrule
\end{tabular}
\label{tab:RC}
\end{table}

\begin{table}[t]
\caption{
F1 represents only using fine-grained type id. F1 with SE represents additionally aggregating semantic representations of fine-grained types. F1 with DLE represents aggregating double-level entity information.}
\footnotesize
\centering
\begin{tabular}{c c c c}
\toprule
{\textbf{Model}} & {\textbf{F1}} & {\textbf{F1 with SE}} & {\textbf{F1 with DLE}} \\ \midrule
BiLSTM & {49.28} & {50.26} & {49.81}\\
GAIN & {58.35} & {58.69} & {58.61}\\
JEREX & {58.19} &{58.26} & {58.34}\\
SSAN & {58.24} & {58.75} & {58.30}\\
\bottomrule
\end{tabular}
\label{tab:DLESE}
\end{table}


\section{Conclusion}
\label{sec:bibtex}

In this paper, we introduce \docredfe{}, a new dataset consisting of double-level entity types, including 11 coarse-grained types and 119 fine-grained types. Compared with existing JERE dataset, \docred{}, our proposed dataset provides a more difficult benchmark for current JERE models and richer entity-level information. To explore the advantages of our new entity schema, we propose and experiment some simple, general but effective ways. We hope \docredfe{} can contribute to future research on building accurate and robust JERE models.

\section{Acknowledgement}

We thank the anonymous reviewers for their helpful comments on this paper. This work was partially supported by National Key Research and Development Project (2022YFC3600402) and National Natural Science Foundation of China (61876009).


\bibliographystyle{IEEEbib}
\bibliography{strings,refs}

\end{document}